\documentclass[runningheads]{llncs}

\usepackage[T1]{fontenc}
\usepackage{lmodern}
\usepackage{graphicx}
\usepackage{amsmath,amssymb,amsfonts}
\usepackage{booktabs}
\usepackage{multirow}
\usepackage{algorithm}
\usepackage{algpseudocode}
\usepackage{xcolor}
\usepackage{float}
\usepackage{pgfplots}
\pgfplotsset{compat=1.18}
\usepackage{wrapfig}
\usepackage{enumitem}
\setlist{nosep}
\usepackage{subcaption}
\usepackage[hidelinks]{hyperref}

\usepackage{orcidlink}
\newcommand{\meanstd}[2]{\ensuremath{#1}\,{\scriptsize\ensuremath{\pm #2}}}

\newcommand{\parag}[1]{\vspace{-3mm}\paragraph{#1}}

\makeatletter
\newcommand{\printfnsymbol}[1]{%
  \textsuperscript{\@fnsymbol{#1}}%
}
\makeatother

\begin{document}

\title{NanoNet: Ultra-Lightweight Network for MRI Brain Tumor Classification toward Edge AI}

\titlerunning{Ultra-Lightweight Network for MRI Brain Tumor Classification}

\author{
Tran Nguyen-Le\inst{1,2}\orcidlink{0009-0009-0736-6244}
\and
Huu-Thuan Huynh\inst{1,2}\orcidlink{0009-0003-3984-3355}
\and
Trung-Nghia Le\inst{1,2}\orcidlink{0000-0002-7363-2610}\thanks{Corresponding author.}
}

\authorrunning{T. Nguyen-Le et al.}

\institute{$^1$ University of Science, Ho Chi Minh City, Vietnam\\
    $^2$ Vietnam National University, Ho Chi Minh City, Vietnam\\
\email{\texttt{nltran2525@apcs.fitus.edu.vn, hhthuan@hcmus.edu.vn, ltnghia@fit.hcmus.edu.vn}}}

\maketitle

\begin{abstract}
    Automated brain tumor classification from magnetic resonance imaging (MRI) can support accessible medical image analysis, but conventional deep learning models often exceed the resource budgets of highly constrained edge-AI platforms. We present NanoNet, an ultra-lightweight convolutional neural network designed for resource-constrained MRI classification and motivated by future edge-AI applications. Inspired by MobileNetV2, NanoNet substantially reduces network depth, channel capacity, expansion ratios, and final feature dimensionality, resulting in only five inverted residual blocks and 8,132 parameters. We further investigate dependency-aware structured pruning, knowledge distillation, and INT8 quantization-aware training for additional compression. Experiments on the BRISC 2025 benchmark show that NanoNet achieves 92.80\% accuracy with 8,132 parameters and 10.498 MFLOPs. Across three compression seeds, it exceeds the mean accuracy of 70\%-pruned MobileNetV2 by 5.20 percentage points while using 60.2\% fewer parameters and 9.8\% less computation. Converted INT8 NanoNet achieves 89.70\% accuracy and across three QAT seeds. Earlier TensorFlow Lite measurements indicate 27.7~KB of baseline model storage, motivating future edge-AI deployment.

\end{abstract}

\section{Introduction}

Brain tumors are challenging to analyze in medical imaging because their appearance varies across tumor types, anatomical locations, imaging protocols, and patients. Magnetic resonance imaging (MRI) provides rich structural and contrast information and is widely used for brain tumor assessment. Automated MRI classification using deep learning has consequently become an active research direction, with CNN-based methods demonstrating strong performance in distinguishing glioma, meningioma, and pituitary tumors~\cite{cheng2015enhanced,deepak2019brain,reza2022cnn}.

Despite their predictive capability, many deep learning models for medical image analysis are designed for computational environments with substantially more resources than highly constrained embedded platforms~\cite{han2015deep}. Their parameter count, computational complexity, and memory requirements can make local inference difficult on resource-limited hardware. Edge AI offers an attractive alternative by enabling computation closer to the data source with reduced dependence on cloud infrastructure~\cite{zhou2019edge}. However, realizing this potential requires models that maintain useful accuracy under extremely small computational, memory, and storage budgets, as emphasized by platform-aware adaptation and tiny-device neural networks~\cite{yang2018netadapt,lin2020mcunet}.

Lightweight architectures such as MobileNet~\cite{howard2017mobilenets} and MobileNetV2~\cite{sandler2018mobilenetv2} reduce computation through depthwise convolutions and inverted residual blocks, while subsequent designs such as ShuffleNet~\cite{zhang2018shufflenet} and MobileNetV3~\cite{howard2019searching} further optimize lightweight networks for mobile and embedded platforms. However, these architectures may still retain unnecessary capacity under extreme resource constraints. Structured pruning and quantization can provide further reductions, although aggressive post-hoc compression may substantially degrade predictive performance. This motivates an architecture-first strategy: instead of starting from a conventional lightweight network and relying primarily on aggressive compression, we design a compact architecture specifically for an extreme resource regime and subsequently investigate additional compression techniques.

To this end, we present NanoNet, an ultra-lightweight CNN for resource-constrained MRI brain tumor classification. Inspired by MobileNetV2, NanoNet substantially reduces architectural capacity by using only eight initial channels, five inverted residual blocks, expansion ratios $t \in \{1,2\}$, and a 48-channel final representation. We further investigate a deployment-oriented compression pipeline combining dependency-aware structured pruning, knowledge-distillation-based fine-tuning, and INT8 quantization-aware training. On the BRISC 2025 benchmark, NanoNet achieves 92.80\% accuracy and 92.97\% macro F1 with only 8,132 parameters and 10.498 MFLOPs, providing a competitive accuracy-efficiency operating point under an extremely constrained computational budget. These results indicate strong potential for future edge-AI implementation.

Our contributions are as follows:

\begin{itemize}
    \item We introduce NanoNet, an ultra-lightweight CNN specifically designed for extremely resource-constrained MRI brain tumor classification. The architecture adopts an architecture-first design that substantially reduces network depth, channel capacity, expansion ratios, and final feature dimensionality relative to MobileNetV2.

    \item We investigate a deployment-oriented compression pipeline that combines dependency-aware structured pruning, knowledge-distillation-based fine-tuning, and INT8 quantization-aware training. We further characterize the resulting models in terms of computational cost, runtime activation memory, and persistent model storage.

    \item We systematically evaluate the accuracy0efficiency trade-off between NanoNet and MobileNetV2 0.1$\times$ under different computational budgets, demonstrating the advantage of architecture-first reduction in the extreme low-computation regime.
\end{itemize}

\section{Related Work}

\parag{\textbf{Deep Learning for Brain Tumor Classification.}}

CNNs have been widely applied to automated brain tumor classification from MRI~\cite{cheng2015enhanced,deepak2019brain,reza2022cnn}. While these methods achieve strong predictive performance, they primarily emphasize classification accuracy and often rely on relatively large CNN architectures, leaving computational and memory efficiency less explored. This motivates studying brain tumor classification under an explicit accuracy--efficiency trade-off, particularly for highly resource-constrained edge-AI platforms.



\parag{\textbf{Lightweight Networks for Edge AI.}}

MobileNet~\cite{howard2017mobilenets} introduced depthwise separable convolutions for efficient mobile vision, while MobileNetV2~\cite{sandler2018mobilenetv2} further improved efficiency through inverted residuals and linear bottlenecks. Subsequent architectures, including ShuffleNet~\cite{zhang2018shufflenet}, MobileNetV3~\cite{howard2019searching}, and GhostNet~\cite{han2020ghostnet}, explored efficient channel operations, architecture search, and inexpensive feature generation for resource-constrained devices. Platform-aware approaches such as NetAdapt~\cite{yang2018netadapt} and MCUNet~\cite{lin2020mcunet} further emphasize deployment constraints beyond conventional FLOPs and parameter counts. These studies motivate our architecture-first strategy, which targets an extremely small computational regime before applying additional compression.

\parag{\textbf{Structured Compression and Quantization.}}

Structured pruning reduces computational cost by removing complete filters or channels rather than individual weights. Representative approaches include filter pruning~\cite{li2016pruning}, network slimming~\cite{liu2017learning}, and ThiNet~\cite{luo2017thinet}, while recent surveys provide a broader analysis of structured pruning methods and their hardware implications~\cite{he2023structured}. Knowledge distillation~\cite{hinton2015distilling} provides an additional mechanism for recovering predictive performance in compact models, and subsequent studies have examined practical factors affecting distillation effectiveness~\cite{beyer2022knowledge}. For low-precision deployment, integer-arithmetic inference and quantization-aware training have been extensively studied~\cite{jacob2018quantization}, with hardware-aware quantization further incorporating deployment constraints into the quantization process~\cite{wang2019haq}.



\section{Proposed Method}

\subsection{Design Objective}

Our goal is to construct a CNN whose resource characteristics make it a promising candidate for future edge-AI applications while maintaining sufficient classification capability. We therefore target three requirements: (1) Low computational cost, to support inference under highly constrained processing budgets; (2) Low model-storage requirements, to support platforms with limited persistent storage; (3) Sufficient classification capacity, to maintain useful performance for four-class MRI classification. 

Rather than treating these requirements solely as post-training optimization objectives, we incorporate them directly into the architecture. This architecture-first design is intended to establish an extremely small computational baseline before applying additional compression. This perspective is consistent with prior work on platform-aware network adaptation~\cite{yang2018netadapt} and tiny-device neural networks~\cite{lin2020mcunet}, which emphasize designing models around stringent deployment constraints rather than relying solely on post-hoc compression.

\begin{table}[t]
\centering
\caption{Per-layer configuration, parameter count (including Batch Normalization), and computational cost of NanoNet at $224 \times 224 \times 3$ input resolution.}
\label{tab:nanonet_layers}
\setlength{\tabcolsep}{6pt}
\resizebox{\linewidth}{!}{%
\begin{tabular}{llccrrr}
\hline
\textbf{Stage} & \textbf{Input} & \textbf{Operator} & $t$ & $s$ & \textbf{\#Params} & \textbf{MACs} \\
\hline
Stem & $224^2 \times 3$ & Conv2d $3\times3$ & -- & 2 & 232 & 3,010,560 \\
Stage 1 & $112^2 \times 8$ & InvertedResidual ($8 \to 8$) & 1 & 2 & 168 & 551,936 \\
Stage 2 & $56^2 \times 8$ & InvertedResidual ($8 \to 16 \to 12$) & 2 & 2 & 552 & 871,808 \\
Stage 3 & $28^2 \times 12$ & InvertedResidual ($12 \to 24 \to 16$) & 2 & 2 & 1,016 & 420,224 \\
Stage 4.1 & $14^2 \times 16$ & InvertedResidual ($16 \to 32 \to 24$) & 2 & 2 & 1,744 & 177,968 \\
Stage 4.2 & $7^2 \times 24$ & InvertedResidual ($24 \to 48 \to 24$) & 2 & 1 & 2,976 & 150,528 \\
Head & $7^2 \times 24$ & Conv2d $1\times1$ ($24 \to 48$) & -- & 1 & 1,248 & 63,504 \\
Classifier & $7^2 \times 48$ & GlobalAvgPool + Linear ($48 \to 4$) & -- & -- & 196 & 2,548 \\
\hline
\textbf{Total} & -- & -- & -- & -- & \textbf{8,132} & \textbf{5,249,076} \\
\hline
\end{tabular}
}
\end{table}

\subsection{NanoNet Architecture}
\label{sec:architecture}

\begin{figure}[t!]
    \centering
    \includegraphics[width=\linewidth, trim={0mm 50mm 0mm 0mm}, clip=true]{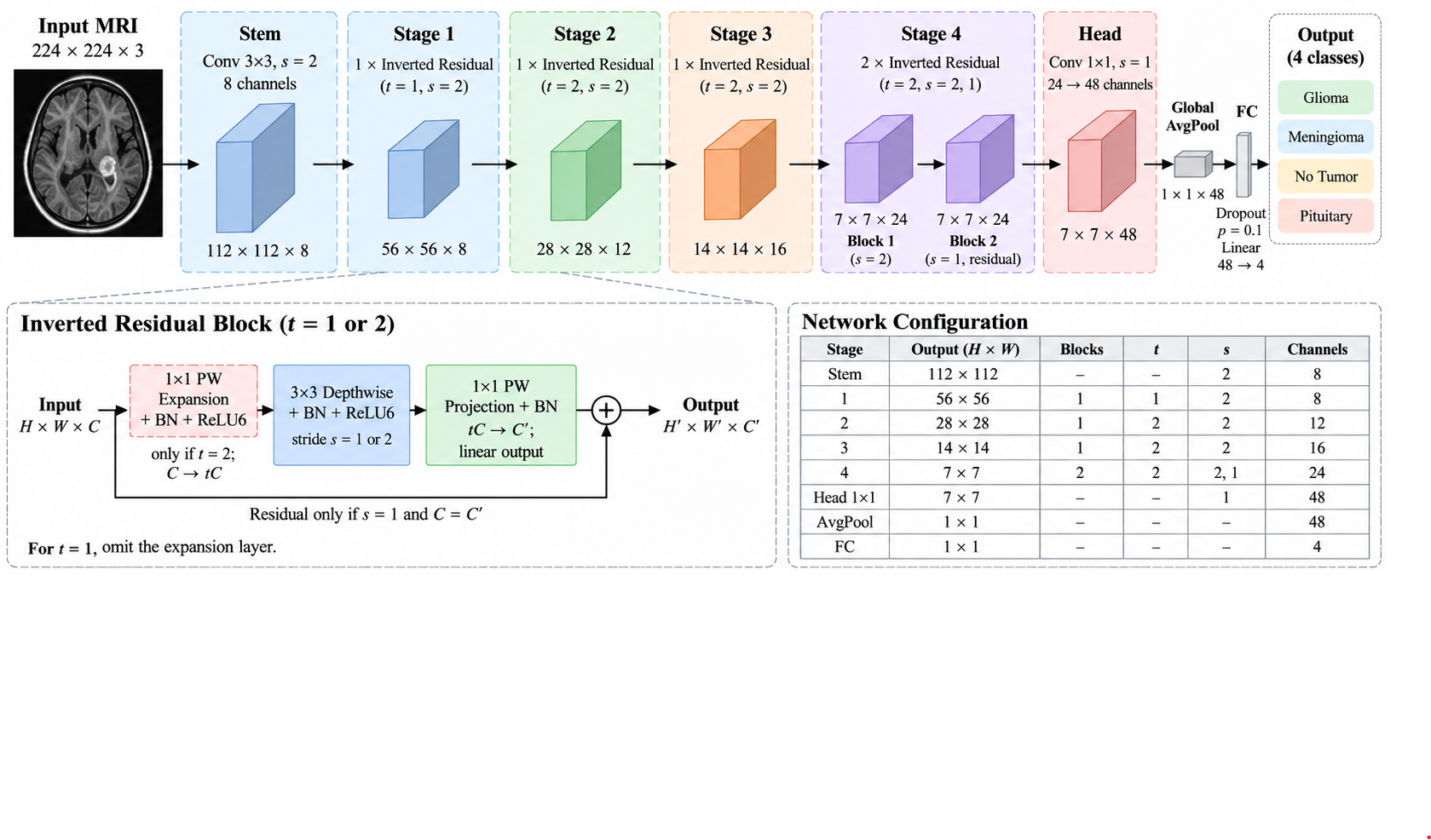}
    \caption{Overview of the NanoNet architecture. Starting from a $224\times224\times3$ MRI image, NanoNet employs a compact convolutional stem followed by five inverted residual blocks organized into four stages. The architecture uses reduced channel capacity and expansion ratios $t\in\{1,2\}$, followed by global average pooling and a four-class classification head.}
    \label{fig:architechture}
    \vspace{-3mm}
\end{figure}

As illustrated in Fig.~\ref{fig:architechture}, NanoNet adopts the inverted residual and depthwise convolution design of MobileNetV2~\cite{sandler2018mobilenetv2}, while substantially reducing architectural capacity for extreme resource constraints. This architecture-first design minimizes capacity before applying further compression:

\begin{itemize}
\item \textit{Initial Channel Capacity Reduction}: The first convolutional layer produces only 8 channels, compared with 32 channels in the reference configuration.
\item \textit{Depth Reduction}: The architecture is reduced from 19 inverted residual blocks across seven stages to five blocks across four stages.
\item \textit{Restricted Expansion Ratios}: To prevent internal parameter explosion, the expansion ratio in each Inverted Residual Block is tightly constrained from the standard $t=6$ to $t \in \{1, 2\}$.
\item \textit{Compact Final Representation}: Instead of expanding the final representation to 1,280 channels, NanoNet uses only 48 channels before classification.
\end{itemize}

\parag{\textbf{Lightweight Inverted Residual Block.}}

NanoNet adopts the inverted residual design of MobileNetV2 as its fundamental building block. Unlike conventional residual blocks that first reduce and then expand the feature dimension, an inverted residual block expands the input channels into a higher-dimensional intermediate representation, applies a depthwise convolution, and subsequently projects the features back to a compact output representation.

Given an input feature map $\mathbf{x} \in \mathbb{R}^{H \times W \times C}$, an inverted residual block with expansion ratio $t$ first applies a $1\times1$ pointwise convolution to expand the channel dimension from $C$ to $tC$. A $3\times3$ depthwise convolution then processes each channel independently, followed by a second $1\times1$ pointwise convolution that projects the representation to the output dimension $C'$. Batch normalization and ReLU6 activations are applied to the intermediate transformations. Formally, the block can be expressed as: 
$
\mathbf{z} =
\phi_{\mathrm{proj}}
\left(
\mathrm{PW}_{\mathrm{proj}}
\left(
\mathrm{DW}_{3\times3}
\left(
\phi_{\mathrm{exp}}
\left(
\mathrm{PW}_{\mathrm{exp}}(\mathbf{x})
\right)
\right)
\right)
\right),
$ 
where $\mathrm{PW}_{\mathrm{exp}}$ and $\mathrm{PW}_{\mathrm{proj}}$ denote the expansion and projection pointwise convolutions, respectively, $\mathrm{DW}_{3\times3}$ denotes the depthwise convolution, and $\phi$ represents the corresponding normalization and activation operations.

A residual connection is employed when the input and output feature maps have identical spatial dimensions and channel numbers. Thus,
\begin{equation}
\mathbf{y} =
\begin{cases}
\mathbf{x} + \mathbf{z}, & \text{if } s=1 \text{ and } C=C',\\
\mathbf{z}, & \text{otherwise},
\end{cases}
\end{equation}
where $s$ denotes the stride of the depthwise convolution.

To make the block suitable for highly constrained edge devices, NanoNet substantially restricts the expansion ratio. While the original MobileNetV2 commonly uses $t=6$, NanoNet limits the expansion ratio to
$t\in\{1,2\}$.
This design reduces the number of intermediate channels and consequently lowers both parameter count and computational cost. Combined with the reduced network depth and channel capacity described in Section~\ref{sec:architecture}, the resulting inverted residual blocks provide sufficient feature transformation capacity while maintaining a compact computational footprint.

\subsection{Deployment-Oriented Compression}

\paragraph{\textbf{Overview.}}

\begin{figure}[t!]
    \centering
    \includegraphics[width=\linewidth]{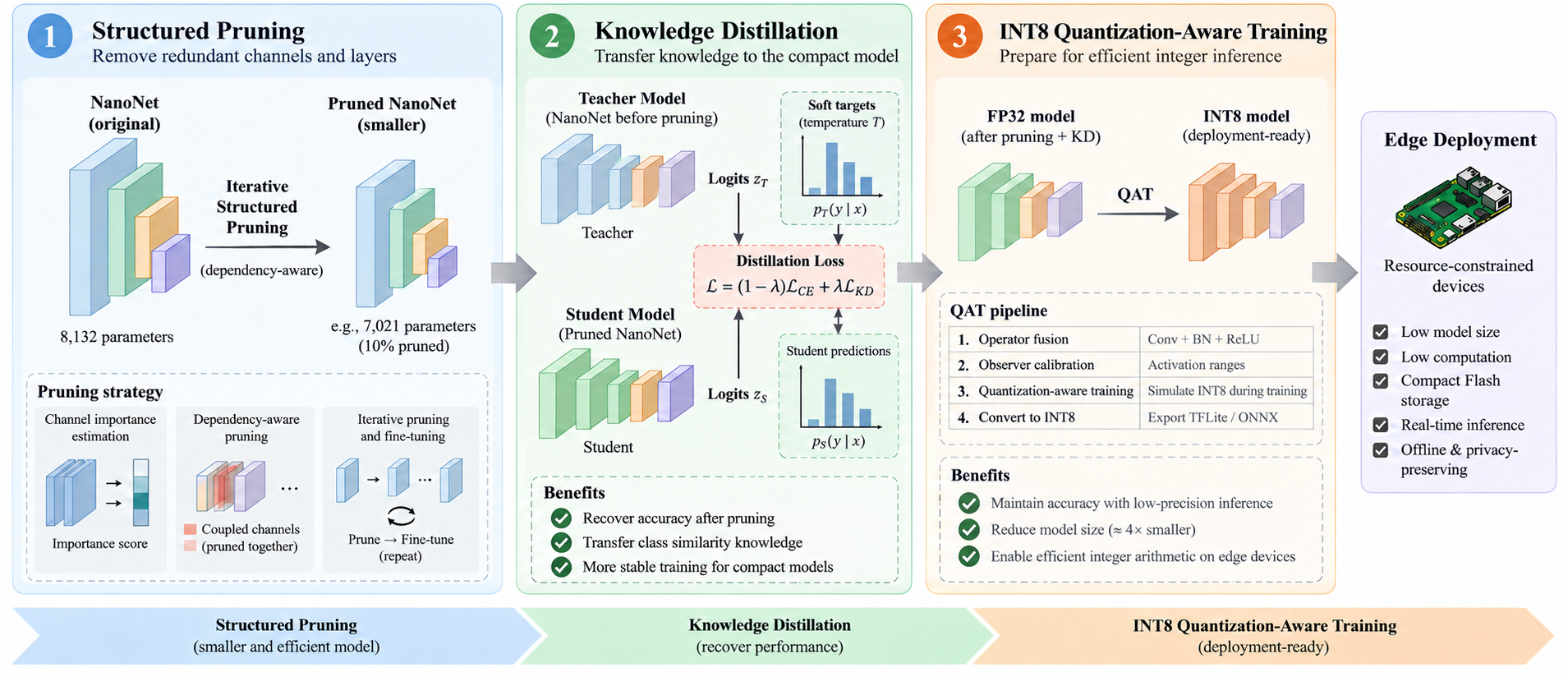}
    \caption{Overview of compression pipeline. The original NanoNet is progressively compressed through dependency-aware structured pruning, knowledge-distillation-based fine-tuning, and INT8 quantization-aware training (QAT), producing a compact model with deployment-oriented resource characteristics.}
    \label{fig:compression_pipeline}
    \vspace{-3mm}
\end{figure}

Although the architecture-first design of NanoNet substantially reduces the computational and parameter budgets, further compression can provide additional resource savings. We therefore investigate a deployment-oriented compression pipeline consisting of structured pruning~\cite{he2023structured,li2016pruning,liu2017learning}, knowledge distillation (KD)~\cite{hinton2015distilling}, and INT8 quantization-aware training (QAT)~\cite{jacob2018quantization}.

The overall pipeline is illustrated in Fig.~\ref{fig:compression_pipeline}. The pre-compression network  is used as a teacher throughout the compression process. The compact model is progressively optimized while undergoing structural pruning and low-precision adaptation. In this way, the original model provides a stable source of predictive knowledge for recovering performance after each compression operation.

\parag{\textbf{Structured Pruning.}}

NanoNet is further compressed using iterative structured pruning. Rather than removing individual weights, which may introduce irregular sparsity with limited practical acceleration, we remove entire coupled channels to obtain a structurally compact network.

We employ the DepGraph framework ~\cite{fang2023depgraph} to identify dependencies between coupled parameters. This is particularly important for inverted residual blocks, where the expansion, depthwise, and projection layers are structurally coupled. Directly pruning a single layer without accounting for these dependencies can lead to incompatible tensor dimensions or invalid residual connections.

Given a target pruning ratio $r$, channel importance is first estimated and the corresponding coupled parameters are grouped according to the dependency graph. The reported pruning ratio denotes the target channel-removal ratio supplied to the dependency-aware pruning procedure rather than the resulting percentage reduction in total parameters. Because coupled layers are pruned jointly, the resulting parameter and computational reductions can differ from the nominal pruning ratio.

The selected groups are then removed simultaneously. The resulting network is subsequently fine-tuned before the next pruning step. This process is repeated iteratively to progressively reduce the model size while preserving the structural integrity of NanoNet.

Formally, let $\mathcal{M}_0$ denote the original NanoNet and
$\mathcal{M}_k$ denote the model after the $k$-th pruning step. The iterative procedure can be summarized as
\begin{equation}
\mathcal{M}_{k+1}
=
\mathcal{P}(\mathcal{M}_k, r_k),
\end{equation}
where $\mathcal{P}(\cdot)$ represents dependency-aware structured pruning and $r_k$ is the pruning ratio at step $k$.
After each pruning operation, the resulting model is fine-tuned using knowledge distillation from the original pre-compression model.

\parag{\textbf{Knowledge Distillation.}}

Aggressive pruning can substantially reduce the representational capacity of a compact network and consequently degrade classification accuracy. To compensate for this degradation, we employ knowledge distillation, in which the original pre-compression NanoNet acts as the teacher and the pruned model acts as the student.

Let $\mathbf{z}_T$ and $\mathbf{z}_S$ denote the logits produced by the teacher and student networks, respectively. The softened class distributions are computed as
\begin{equation}
p_T(y\mid x)
=
\mathrm{softmax}
\left(
\frac{\mathbf{z}_T}{T}
\right),
\qquad
p_S(y\mid x)
=
\mathrm{softmax}
\left(
\frac{\mathbf{z}_S}{T}
\right),
\end{equation}
where $T$ is the temperature parameter.

The student is optimized using a combination of the standard classification loss and the distillation loss:
\begin{equation}
\mathcal{L}
=
(1-\lambda)\mathcal{L}_{CE}
+
\lambda\mathcal{L}_{KD},
\end{equation}
where $\mathcal{L}_{CE}$ is the cross-entropy loss, $\mathcal{L}_{KD}$ measures the discrepancy between the teacher and student predictions, and $\lambda$ controls the contribution of knowledge distillation. 

In our compression pipeline, KD is applied after each pruning step during fine-tuning. Consequently, the teacher provides a stable supervisory signal while the student progressively adapts to its reduced architecture. This compression-aware strategy helps recover classification performance without increasing the inference-time complexity of the final model.

\parag{\textbf{Quantization-Aware Training.}}

After structural pruning and knowledge-distillation-based fine-tuning, the resulting FP32 model is converted into an INT8 model for efficient edge inference. We employ quantization-aware training (QAT)~\cite{jacob2018quantization} to expose the network to quantization effects during training and reduce the accuracy degradation caused by low-precision arithmetic.

Before QAT, consecutive convolution, batch normalization, and activation operations are fused where applicable. Residual connections are also adapted to ensure compatibility with integer arithmetic. Activation observers are subsequently calibrated using a small subset of the training data to estimate appropriate activation ranges.

During QAT, fake-quantization operations simulate INT8 rounding and saturation while the underlying parameters remain trainable. The model therefore learns to compensate for the numerical errors introduced by low-precision representation. 

After QAT, the trained network is converted to an 8-bit integer representation suitable for deployment-oriented evaluation. The resulting model provides a low-storage integer representation that can facilitate future implementation on resource-constrained edge-AI hardware.






\section{Experiments}

\subsection{Experimental Settings}

\paragraph{\textbf{Dataset.}}
We evaluate NanoNet on the BRISC 2025 dataset~~\cite{Fateh_2026}, which contains 6,000 T1-weighted brain MRI images from four classes: Glioma, Meningioma, Pituitary Tumor, and No Tumor. Following the provided benchmark split, 5,000 images are used for training and 1,000 for testing.

\parag{\textbf{Three-Seed Protocol.}}
Each configuration is evaluated with seeds 42, 43, and 44. Every run starts from an independent copy of the same original pretrained checkpoint for its model family, rather than from another pruning-ratio checkpoint. Thus, these are three compression/QAT seeds, not three independently trained baseline models. The uncompressed FP32 baseline is the same fixed pretrained model evaluated three times, so its standard deviation is zero. The original teacher-training split is undocumented; the validation holdout applies to compression/QAT and cannot establish that these images were unseen during teacher training.

\parag{\textbf{Evaluation Metrics.}}
We evaluate models in terms of classification accuracy and resource efficiency. Accuracy is measured on the test set for both FP32 models and their corresponding INT8 versions. Resource efficiency is characterized by the number of trainable parameters and computational complexity in GFLOPs at $224\times224$ input resolution. For the INT8 models, we additionally report peak TensorFlow Lite Tensor Arena memory during single-batch inference and persistent model storage measured by the size of the converted \texttt{.tflite} model. These measurements characterize deployment-oriented resource requirements rather than performance on a physical edge device.

\parag{\textbf{Baseline.}}
We use MobileNetV2 0.1$\times$~~\cite{sandler2018mobilenetv2} as the primary baseline because it follows the same inverted residual design philosophy as NanoNet while providing a substantially larger lightweight architecture. To evaluate the benefit of architecture-first reduction under extreme resource constraints, we compare NanoNet with MobileNetV2 under multiple structured-pruning ratios rather than considering only the uncompressed baseline.

\parag{\textbf{Implementation Details.}}
All models are implemented in PyTorch and trained and evaluated using dual NVIDIA Tesla T4 GPUs with an Intel x86-64 CPU. All experiments use an input resolution of $224\times224$. Parameter counts and computational complexity are measured using the torch-pruning profiling utility. For the compression experiments, models are first evaluated after structured pruning and knowledge-distillation-based fine-tuning, followed by INT8 conversion through QAT. QAT uses a learning rate of $5\times10^{-5}$ before the final INT8 conversion.

\subsection{Results}

\begin{table}[t!]
\centering
\caption{FP32 classification performance and computational cost over three seeds.}
\label{tab:main-results}
\label{tab:efficiency}
\small
\setlength{\tabcolsep}{5pt}
\begin{tabular}{lrrrr}
\toprule
\textbf{Model} & \textbf{Params} & \textbf{MFLOPs} & \textbf{Accuracy (\%)} & \textbf{Macro F1 (\%)} \\
\midrule
\multicolumn{5}{l}{\textbf{MobileNetV2 0.1$\times$}} \\
Baseline & 91,204 & 62.956 & \meanstd{97.80}{0.00} & \meanstd{97.81}{0.00} \\
30\% pruned & 60,564 & 37.949 & \meanstd{97.23}{0.76} & \meanstd{97.42}{0.74} \\
50\% pruned & 40,145 & 25.660 & \meanstd{96.13}{0.31} & \meanstd{96.25}{0.38} \\
60\% pruned & 30,069 & 18.113 & \meanstd{93.60}{0.20} & \meanstd{93.78}{0.10} \\
70\% pruned & 20,449 & 11.637 & \meanstd{87.60}{2.80} & \meanstd{87.86}{2.72} \\
75\% pruned & 14,994 & 11.076 & \meanstd{82.57}{3.57} & \meanstd{83.08}{3.80} \\
90\% pruned & 3,773 & 3.648 & \meanstd{62.00}{7.18} & \meanstd{60.79}{8.45} \\
\midrule
\multicolumn{5}{l}{\textbf{NanoNet}} \\
Baseline & 8,132 & 10.498 & \meanstd{92.80}{0.00} & \meanstd{92.97}{0.00} \\
10\% pruned & 7,021 & 7.854 & \meanstd{89.23}{2.76} & \meanstd{89.34}{2.72} \\
25\% pruned & 5,280 & 3.604 & \meanstd{75.67}{2.59} & \meanstd{75.76}{2.69} \\
\bottomrule
\end{tabular}
\end{table}

\paragraph{\textbf{Classification Performance and Computational Efficiency.}}

Table~\ref{tab:efficiency} compares MobileNetV2 0.1$\times$ and NanoNet under different structured-pruning configurations. The results reveal a clear trade-off between computational reduction and classification performance.

The uncompressed MobileNetV2 baseline achieves 97.80\% accuracy and 97.81\% macro F1 with 91,204 parameters and 62.956 MFLOPs. NanoNet achieves 92.80\% accuracy and 92.97\% macro F1 with 8,132 parameters and 10.498 MFLOPs, reducing parameters by 91.1\% and computational cost by 83.3\% relative to that baseline.

More importantly, the comparison becomes particularly relevant in the extreme low-computation regime. The 70\%-pruned MobileNetV2 achieves 87.60\% accuracy and 87.86\% macro F1 at a mean cost of 11.637 MFLOPs and 20,448.7 parameters. NanoNet therefore has 5.20 percentage points higher mean accuracy and 5.11 points higher mean macro F1 while using 60.2\% fewer parameters and 9.8\% less computation.

This result does not indicate that NanoNet is universally more accurate than MobileNetV2. Rather, it shows that architecture-first reduction can provide a more favorable accuracy-efficiency operating point when the computational budget is extremely constrained.


\parag{\textbf{Accuracy-Efficiency Trade-off.}}

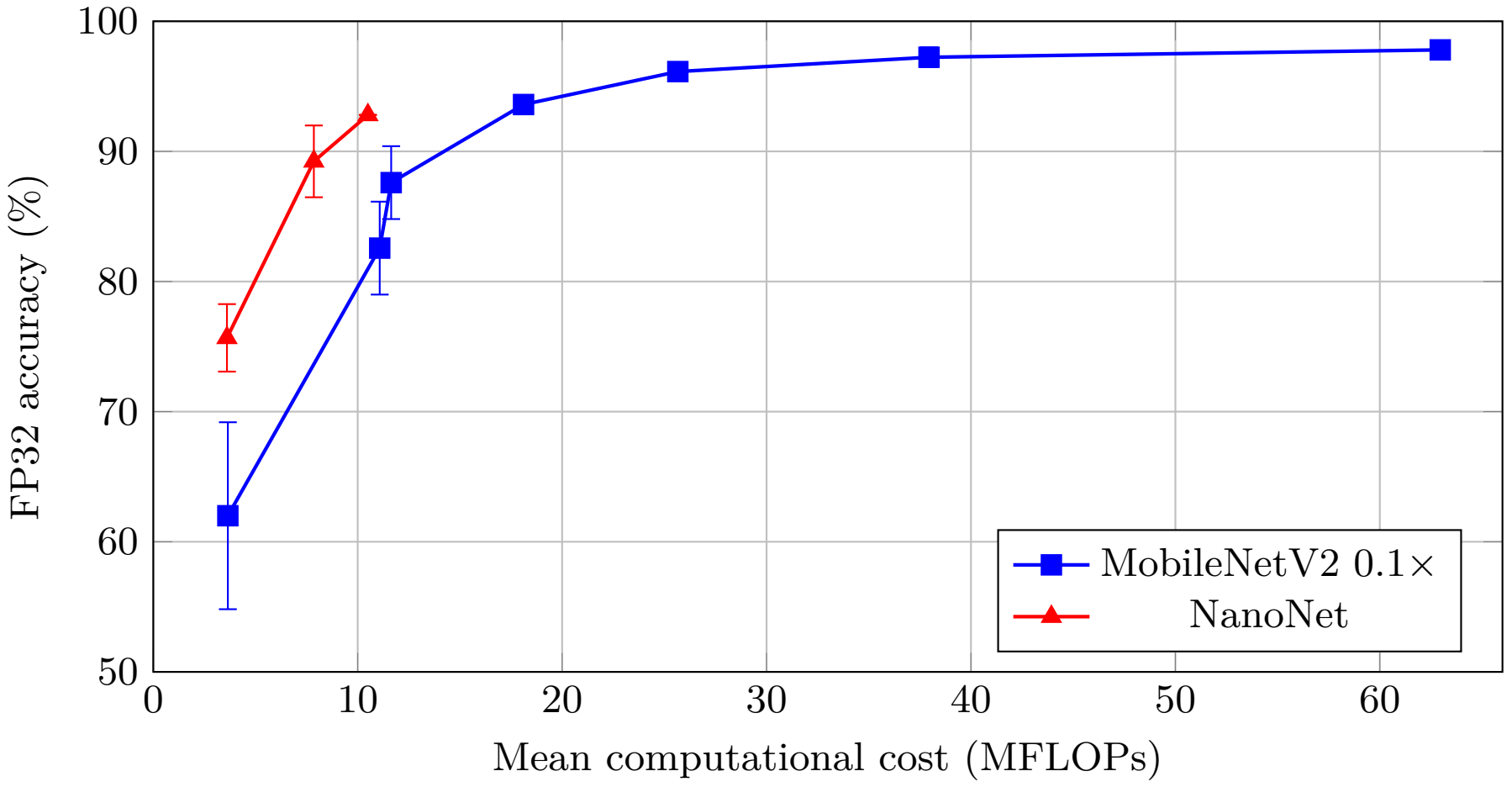
\begin{figure}[t!]
\centering
\begin{tikzpicture}
\begin{axis}[
width=\linewidth,height=6.7cm,
xlabel={Mean computational cost (MFLOPs)},
ylabel={FP32 accuracy (\%)},
grid=major,legend pos=south east,mark size=2pt,
xmin=0,xmax=66,ymin=50,ymax=100,
legend style={font=\small},
]
\addplot[color=blue,mark=square*,thick,error bars/.cd,y dir=both,y explicit] coordinates {
(3.647932,62.0000) +- (0,7.1840)
(11.076301,82.5667) +- (0,3.5698)
(11.637149,87.6000) +- (0,2.8000)
(18.113068,93.6000) +- (0,0.2000)
(25.659781,96.1333) +- (0,0.3055)
(37.948746,97.2333) +- (0,0.7638)
(62.956040,97.8000) +- (0,0.0000)
};
\addlegendentry{MobileNetV2 0.1$\times$}
\addplot[color=red,mark=triangle*,thick,error bars/.cd,y dir=both,y explicit] coordinates {
(3.604000,75.6700) +- (0,2.5929)
(7.854322,89.2333) +- (0,2.7592)
(10.498152,92.8000) +- (0,0.0000)
};
\addlegendentry{NanoNet}
\end{axis}
\end{tikzpicture}
\caption{FP32 accuracy versus computational cost. Points show three-seed means; vertical bars indicate $\pm1$ sample std.}
\label{fig:tradeoff}
\end{figure}

Fig.~\ref{fig:tradeoff} plots mean FP32 accuracy against mean MFLOPs, with error bars showing the sample standard deviation of accuracy. MobileNetV2 maintains high accuracy at relatively larger computational budgets, but its performance decreases substantially under aggressive pruning.

NanoNet operates in a substantially smaller computational regime. Its baseline configuration achieves 92.8\% accuracy at only 10.498 MFLOPs, whereas further pruning to 7.854 and 3.604 MFLOPs results in 89.23\% and 75.67\% accuracy, respectively.

These results indicate that architecture-first reduction can provide a more favorable operating point than applying aggressive pruning to a relatively larger lightweight architecture. In particular, NanoNet achieves substantially higher accuracy than heavily pruned MobileNetV2 at comparable computational complexity.

\begin{table}[t!]
\centering
\caption{INT8 performance after QAT. $\Delta$ denotes the change in mean accuracy from FP32 (percentage points).}
\label{tab:quantization}
\small
\setlength{\tabcolsep}{6pt}
\begin{tabular}{lrrr}
\toprule
\textbf{Model} & \textbf{Accuracy (\%)} & \textbf{Macro F1 (\%)} & $\boldsymbol{\Delta}$ \\
\midrule
\multicolumn{4}{l}{\textbf{MobileNetV2 0.1$\times$}} \\
Baseline & \meanstd{97.97}{0.15} & \meanstd{98.05}{0.18} & $+0.17$ \\
30\% pruned & \meanstd{97.13}{1.00} & \meanstd{97.28}{0.92} & $-0.10$ \\
50\% pruned & \meanstd{96.20}{0.20} & \meanstd{96.31}{0.15} & $+0.07$ \\
60\% pruned & \meanstd{93.63}{0.55} & \meanstd{93.92}{0.58} & $+0.03$ \\
70\% pruned & \meanstd{87.27}{2.97} & \meanstd{87.52}{2.96} & $-0.33$ \\
75\% pruned & \meanstd{81.87}{2.85} & \meanstd{82.27}{3.10} & $-0.70$ \\
90\% pruned & \meanstd{60.53}{7.26} & \meanstd{59.13}{8.43} & $-1.47$ \\
\midrule
\multicolumn{4}{l}{\textbf{NanoNet}} \\
Baseline & \meanstd{89.70}{3.41} & \meanstd{90.22}{3.27} & $-3.10$ \\
10\% pruned & \meanstd{89.20}{2.36} & \meanstd{89.47}{2.29} & $-0.03$ \\
25\% pruned & \meanstd{75.70}{2.52} & \meanstd{76.12}{2.26} & $+0.03$ \\
\bottomrule
\end{tabular}
\end{table}


\parag{\textbf{INT8 Quantization.}}

Table~\ref{tab:quantization} compares FP32 and INT8 accuracy after QAT. Quantization affects the two architectures differently. MobileNetV2 exhibits relatively small changes across the evaluated configurations, whereas NanoNet is more sensitive to INT8 conversion because its representation is already highly compact.

For NanoNet, the baseline mean accuracy decreases from 92.80\% in FP32 to 89.70\% after INT8 conversion (with 90.22\% macro F1), corresponding to a 3.10-percentage-point reduction. Meanwhile, the 10\%- and 25\%-pruned configurations show almost no change in mean accuracy, shifting by $-0.03$ (from 89.23\% to 89.20\%) and $+0.03$ percentage points (from 75.67\% to 75.70\%), respectively.

These results show that low-precision conversion can introduce non-negligible accuracy changes for extremely compact networks. Nevertheless, the resulting INT8 models provide substantially smaller integer representations, supporting the potential of NanoNet for future resource-constrained edge-AI implementations.


\parag{\textbf{Memory and Model Storage.}}

The following provides a deployment-oriented analysis of the selected INT8 models. The baseline NanoNet requires only 27.7~KB of model storage after INT8 conversion, while the 10\%-pruned NanoNet requires 27.6~KB. In comparison, the 70\%- and 60\%-pruned MobileNetV2 models require 81.4~KB and 103.0~KB, respectively.

The runtime activation memory exhibits a different behavior. NanoNet requires 394.6~KB of peak Tensor Arena memory in the baseline configuration and 382.4~KB after 10\% pruning, compared with 333.4~KB for the selected MobileNetV2 configurations. This result highlights an important distinction between persistent model storage and runtime activation memory: reducing parameter count and model-file size does not necessarily minimize peak activation memory.

Although these measurements are not obtained on a physical edge device, the extremely small computational footprint and model-storage requirements demonstrate that NanoNet has strong potential for future implementation on highly constrained edge-AI platforms.

\subsection{Analysis of Resource-Accuracy Trade-offs}

\paragraph{\textbf{Pruning-Ratio Analysis.}}

MobileNetV2 maintains relatively high accuracy under moderate pruning but degrades rapidly under aggressive pruning. NanoNet is more sensitive to additional pruning: declining from 92.80\% baseline accuracy to 89.23\% and 75.67\% mean accuracy at 10\% and 25\% pruning. This behavior indicates that the architecture-first design leaves limited redundancy for further structural removal.

\parag{\textbf{Compression Pipeline Analysis.}}

The compression stages address complementary resource constraints. Structured pruning reduces channel capacity and consequently lowers parameter count and computational cost, while knowledge distillation (KD) provides supervisory information during post-pruning fine-tuning. INT8 quantization-aware training (QAT) subsequently adapts the compact FP32 models to low-precision inference and enables a smaller integer representation.

For NanoNet, additional pruning reduces the computational cost from 10.50 MFLOPs to 7.85 and 3.60 MFLOPs at 10\% and 25\% nominal pruning ratios, respectively. However, mean accuracy decreases from 92.80\% to 89.23\% and 75.67\%, indicating that the compact baseline contains limited redundancy and is sensitive to further structural reduction. After QAT, the baseline NanoNet experiences a 3.10\% accuracy reduction (to 89.70\%), whereas the 10\%- and 25\%-pruned models remain virtually unchanged at 89.20\% and 75.70\%, respectively.

These results highlight the complementary roles of the compression stages while also revealing the limitations of aggressive compression for an already compact architecture. Since the current experiments do not include a separate no-KD ablation, the individual contribution of KD cannot be isolated. Nevertheless, the complete pipeline produces highly compact models, with the baseline NanoNet requiring only 27.7~KB of INT8 model storage.

\parag{\textbf{Resource-Budget Analysis.}}

The evaluated models span computational budgets from 62.96 MFLOPs for the original MobileNetV2 0.1$\times$ down to 3.60 MFLOPs for the most aggressively compressed configurations. The comparison near the 11-MFLOP regime is particularly informative: NanoNet achieves 92.80\% accuracy with 8,132 parameters (10.50 MFLOPs), whereas 70\%-pruned MobileNetV2 achieves 87.60\% accuracy with 20,449 parameters (11.64 MFLOPs). Thus, NanoNet provides a 5.20-percentage-point accuracy improvement while using 60.2\% fewer parameters and 9.8\% less computation.

NanoNet's advantage is specific to the evaluated low-resource regime: the larger uncompressed MobileNetV2 remains more accurate. These observations support architecture-first reduction under stringent budgets, but do not establish a universal advantage over all alternatives. A scratch-trained, matched-cost baseline and direct hardware measurements remain necessary to strengthen the deployment comparison.




\subsection{Limitations}

Despite its compact design, NanoNet has four main limitations. First, evaluation on a single benchmark does not establish cross-dataset generalization. Second, the experiments use 2D MRI images rather than volumetric 3D data. Third, the small architectural capacity leaves limited redundancy, making further pruning highly sensitive. Finally, we do not evaluate NanoNet on physical ARM, mobile, or embedded hardware; thus, latency, throughput, power, and energy measurements remain future work.





\section{Conclusion}

We presented NanoNet, an ultra-lightweight CNN for resource-constrained MRI brain tumor classification. Its architecture-first design achieves 92.80\% accuracy and 92.97\% macro F1 with 8,132 parameters and 10.498 MFLOPs. Across three compression seeds, 70\%-pruned MobileNetV2 achieves 87.60\% accuracy at a higher mean cost of 11.637 MFLOPs. NanoNet therefore provides 5.20 percentage points higher mean accuracy with 60.2\% fewer parameters and 9.8\% less computation. Converted INT8 NanoNet achieves 89.70\% accuracy and 90.22\% macro F1. These repeated runs quantify compression/QAT variability for fixed pretrained baselines. Earlier single-run TensorFlow Lite storage measurements further motivate deployment, while independent baseline retraining, matched-cost comparisons, physical edge-device evaluation, and extension to volumetric MRI remain future work.



\begin{credits}
\subsubsection{\ackname} 
This research is funded by Vietnam National University - Ho Chi Minh City (VNU-HCM) under Grant Number A2025-18-04.
\end{credits}

\bibliographystyle{splncs04}
\bibliography{references}

\end{document}